# On Analyzing Churn Prediction in Mobile Games

KIHOON JANG

Computer Science and Information Technology, University of the District of Columbia, Washington, DC. USA

JUNWHAN KIM

Computer Science and Information Technology, University of the District of Columbia, Washington, DC. USA

BYUNGGU YU

Computer Science and Information Technology, University of the District of Columbia, Washington, DC. USA

In subscription-based businesses, the churn rate refers to the percentage of customers who discontinue their subscriptions within a given time period. Particularly, in the mobile games industry, the churn rate is often pronounced due to the high competition and cost in customer acquisition; therefore, the process of minimizing the churn rate is crucial. This needs churn prediction, predicting users who will be churning within a given time period. Accurate churn prediction can enable the businesses to devise and engage strategic remediations to maintain a low churn rate. The paper presents our highly accurate churn prediction method. We designed this method to take into account each individual user's distinct usage period in churn prediction. As presented in the paper, this approach was able to achieve 96.6\% churn prediction accuracy on a real game business. In addition, the paper shows that other existing churn prediction algorithms are improved in prediction accuracy when this method is applied.

## 1 INTRODUCTION

In customer-based business, the rate of customer churn is one of important issues. The high churn rate of the customers can significantly cause the reduction of the revenue. Many companies have offered a variety of promotions to minimize the churn of customers because the cost for acquiring a new customer is usually higher than retaining old one. If the customers who are likely to churn are detected, we can easily identify the target of promotion. The importance of churn prediction is increasing because it can greatly affect the change of actual profit, the number of customers, and the efficiency of promotion.

A churn prediction is more important for game companies relying on a customer churn rate. It is because playing games cannot take precedence over an essential part of our life. Unlike other customer-based industries, it is easy to start developing and close games. As a game-ecosystem, gamers can easily switch one to another game. This phenomenon is commonly shown in mobile rather than stand-alone games.

Having the ability to accurately predict user's churn is necessary because the prediction's result helps business in gaining a better understanding of expected revenue. The game companies focus on holding many loyal gamers promising greater revenue consistency. The top 1% of game user-accounts produces 20% of total revenue, and the top 10% of game user-accounts produces 50% of total revenue [1]. The number of the users who pay for game items or subscription fees need to be maintained with low churn rate. In addition, the result of churn prediction represents the potential churn rate of a particular user. If the user is expected to churn, the game companies may provide discounted subscription fees and game items with the user. The game companies are more sensitive to the user's churn problem [1]. Thus, retaining existing users is more significant to manage the companies.

Many researches for the churn prediction of game users have been studied [1, 2, 3, 4], and the accuracy of churn prediction has been emphasized. We focus on the following two aspects. First, this paper considers how to exploit the prediction results. Rather than identifying whether or not users will churn, this paper underlines when the users will churn. This is directly related to how to avoid user churn. Second, this paper aims to increase the accuracy of churn prediction. Like the prior studies, the result of our churn prediction can be used by game companies to strategically design various promotions to improve the user retention rates.

Unlike other day-based approaches, the approach presented in this paper is a novel concept – "churn vector". Our experimental results show that our churn-vector-based approach is substantially more accurate in predicting churns in mobile games. Furthermore, this approach enables the campaign manager of a mobile game to design an effective promotion for each user based on the user's churn-vector.

We conducted a set of experiments in churn prediction both using the existing day-based method and using our novel churn-vector-based method. For the prediction algorithm, we tested Least Absolute Shrinkage and Selection Operator (LASSO) [5, 6], Support Vector Machines (SVM) [1, 7], deep neural networks (NNs) [1], Decision Tree (DT) [1, 2, 3], Random Forest (RF) [2, 3], and Gradient Boosted Machine (GBM) [3]. First of all, our churn-vector method shows a better accuracy in predicting churn classifications (classification accuracy) and a smaller prediction error in predicting churn dates (regression error) regardless of the underlying prediction algorithm. In more detail, our churn-vector method equipped with NN (Multi-Layer Perceptron) performed best in both classification and regression.

Since we obtained the best results from the model using NNs in the experiments described in the previous paragraph, we conducted additional experiments to understand which neural networks on the churn prediction in mobile games outperform. We further tested Convolutional Neural Networks (CNN) [3, 4], Long Short-Term Memory networks (LSTM) [3, 8], and Attention Networks (AN) [9] in this experiment. The results of this experiment show the highest accuracy in using an AN. The main contribution of this paper shows the accuracy of our vector-based model on AN is 0.966, which is the most accurate result to the best of our knowledge.

The rest of the paper is organized as follow. We overview related works in Section 2 and our churn prediction model in Section 3. We describe experiment and result in Section 4. We conclude in Section 5.

## 2  RELATED WORK

Recently, various studies have been conducted to predict and analyze the churn of game users. These studies define churn in a variety of ways and use many of the game's actual data sets to predict churns. Julian et al. compared the prediction performance of four common classification algorithms and presented that NNs achieved the best prediction performance in terms of area under curve -- up to 0.93 [1]. Anping *et al*. proposed LASSO for Radial Basis Function Neural Network called L-RBF and achieved up to 80.95\% accuracy [5]. These predict whether a user will churn or not within the week following that day. However, the prior researches are limited to traditional NNs, and the prediction is based on churn day.

In a study conducted in 2017, Seungwook *et al*. tested churn prediction models using play log data from three different genres of games [3]. The genres of the games used in the study were casual action, casual racing, and capture the fag games, and the prediction algorithms performed in this study included logistic regression, gradient boosting, random forests, CNN, and LSTM [3]. The best-performing algorithms were LSTM for casual action games, gradient boosting for casual racing, and logistic regression for capture the flag. The most effective algorithm is different according to the game genre. The best accuracy values were found by tuning the thresholds, and for gradient boosting they turn out to be 0.93, 0.86, and 0.85 for each game. We evaluated churn prediction using the same prediction models with the proposed vector-based churn data to show higher accuracy.

JiHoon *et al*. extracted seven psychological features and payment data from 45 types of raw data logs to improve churn prediction [2]. As a result, this research obtained about 5% more improvement in the model that extracted psychological features than simply using raw log data. Our vector-based churn prediction was shown better results by finding user characteristics through raw log data.

Pei *et al*. compared the performance of deep learning and parametric models to predict the user's lifetime value [4]. This research exploits CNNs to predict the in-app purchases with data from a mobile role-playing game and reduced the computation time of CNN with captured large size video data sets, resulting in low error rate -- 3.96%. However, exploiting other neural networks remained for future works and only lifetime value was estimated. Instead of game screens, our data sets are based on game's log data. The log data is switched to an image for trading.

The difference between churn day and vector is discussed in Section 4. The paper compared our vector-based model with existing churn prediction models and presented how much the accuracy of churn prediction is improved.

## 3 CHURN VECTOR-BASED MODEL

Developing a neural network is to determine the type of problem that needs to be solved. We consider existing neural networks with mobile game data. In this section, we briefly review existing prediction algorithms and present experiment with the algorithms to see how neural networks affect mobile game users' departure prediction. As the traditional prediction algorithms with the best performance, LASSO, SVM, DT, RF, and GBM are discussed. As deep learning approaches, the following prediction algorithms are presented: Feed Forward, CNN, RNN, LSTM and AN.

The algorithms for neural networks are mathematical structures with several processing layers that can separate the features of data. Deep neural networks are typically called feed forward networks in data flows from the input layer to the output layer. However, feed forward networks do not know a sequence of inputs. RNN addresses this issue with a time twist. The output of previous time step is given as input of the next time step in each neuron of RNN. This makes RNN know the sequence. However, RNN also has problems like vanishing gradient or long-term dependency problem where information rapidly gets lost over time. To resolve this issue, LSTM, as a type of RNN, use special units including a memory cell that can maintain information in memory for long periods of time [8, 10]. In the meanwhile, CNN are a class of deep neural networks which is most commonly applied to analyzing visual imagery.

A neural network is a kind of a function approximator. The ability of the approximator depends on its architecture. Attention is a mechanism to improve the performance of encoder-decoder architectures as neural attention [9]. In this paper, the encoder LSTM is used to process the entire input and encode it into a context vector. The attention mechanism to overcome the limitation that allows the network to learn where to pay attention in the input sequence for each item in the output sequence.

To retain current users, game companies need to understand why and when users will churn. However, it is difficult for them to predict exact churn reasons and dates. Instead, whether certain users are expected to churn within a date and how many days users stay before churning can be predicted. For example, given a prediction such that game users will churn in 30 days, it is difficult to infer churn reasons from this prediction. In fact, such a prediction may not help minimizing user attrition because the game company may not build promotion strategies fitting into users expected to churn. We define a churn-vector to achieve high prediction accuracy and plan strategic promotions to minimize game user attrition. The churn-vector can be obtained from the following equation:

$$\text{Churn Vector} = \frac{RemainDays}{\text{Last Play Day} - \text{First Play Day}}$$

Equation 1. Definition of Churn Vector

As shown in Equation (1), the churn-vector is the normalized number of days remaining to the total number of days played by a user. Based on the churn-vector, a campaign manager can effectively apply a promotion to the users expected to churn with the same churn-vector. Multiple user groups can be categorized with churn-vectors, and different promotion strategies can be applied for each group. Moreover, predicting the churn-vector is more accurate than doing churn dates. We compared the accuracies with multiple models and algorithms in Section 4.

Figure 1 shows the play time of two users -- A and B to understand the churn-vector. The play time of users A and B are extract from real game user data. User B's game account holding time is shorter than that of user A. User A is expected to churn after 12 days in Figure 1 (a) because user A's play time is significantly reduced. In this case, a game company assumes that user A may be not interested in playing the game. At the same time, user B is expected to churn, but the play time is not changed significantly. If a promotion package is conducted for both users at this time, this promotion may not be effective for one or both because two users may have different reasons to churn. The churn-day approach cannot identify two users expected to churn. However, the churn-vector produces the levels of churn. Users A and B have different churn-vectors as shown in Figure 1 (b). User A's churn-vector is 30%, and user B's churn vector is 50%, implying that two users have different reasons to churn.

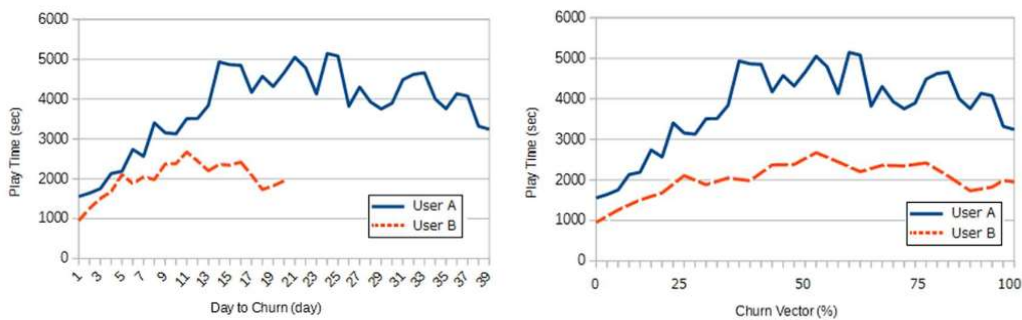

(a) Play Time of Two Users on Churn Day　　(b) Play Time of Two Users on Churn Vector

Figure 1. Difference of Churn-Vector and Churn-Day

As another advantage of using churn-vector, the accuracy of prediction of the churn-vector is higher than that of churn-day. Most users have joined the game at different time and play-patterns. If 20 days are set as maximum days to predict churn-day. As shown in Figure 1 (a), user B played the game for only 20 days. That is why churn-day-based prediction's accuracy is low. The churn-vector is designed to affect evenly the overall data set, increasing its accuracy.

## 4 EXPERIMENTS AND RESULTS

### 4.1 Experiments Setting

As described in the related works, churn prediction is divided into a regression model that predicts the date remaining until churn and a classification model that predicts whether the user will churn. We will experiment with the following four models – Models A and B are based on Regression and $R^2$ score, and Models C and D are based on Classification and Accuracy. As inputs, churn vectors are exploited for Models A and C. and churn days are exploited for Models B and D.

We compared the $R^2$ values of model A with the churn vector and model B with the churn date in the regression model to see how well we can explain the phenomenon when using the churn vector. In the classification model, we compared the accuracy of model C on the churn vector and model D on the churn date to find how the performance improves when the churn vector is used.

### 4.2 Data Set

To collect user data, we chose an idle game – "Maze X Brave" which has been in service for eight months. When a user starts a game, he/she chooses a character and levels-up it with some game resources. The character searches a target on a labyrinth to acquire game resources. The user can decide whether or not to proceed with further character enhancement or to start a new game with more powerful characters. To build the data set of the idle game, we collect data from 800 real users for eight months and predict user's retention for last five months. Due to sufficient data collection, the prediction is not performed for the first three months.

The log data of this game consists of login, stage clear, resource acquisition, and character acquisition. We extract user data according to four steps as shown in Table 1. As the primary player data, the collected data include the date of the primary player data, the date of final access, the day remaining until churn, ranking, total play time, the highest clear stage, total resource acquisition amount, and total character acquisition amount. Based on the primary player data, we extract the secondary data -- play time change amount, daily ranking change amount, clear state change amount, and additional character acquisition amount. With the secondary change data, we extract the data of change rate -- play time change rate, daily ranking change rate, clear state change rate, and additional character acquisition rate.

Table 1. Procedure of Data Extraction

| Steps | Data |
|---|---|
| 1. Log Data | Login Log, Stage Clear Log, Get Resource Log, Get Character Log |
| 2. Player Data of the Day | Final Login Day, Total Play Time, Day Remaining to Churn Ranking, The Highest Clear Stage, Total Resource, Total Character |
| 3. Change of Data | Changes of Ranking, Changes of Play Time, Changes of Clear Stage, Changes of Resource, Changes of Character |
| 4. Change of Rate | Change Rate of Ranking, Change Rate of Play Time, Change Rate of Stage, Change Rate of Resource, Change Rate of Character |

Table 2 presents a list of factors extracted from the user play log data. With these factors, the user's behavior, the user's highest record, the user's change, and the change in the user's sense of achievement can be derived. Thus, we categorized these factors into the above four types of information. Upon the experiment, we investigated which information has the most influence on the churn prediction.

Table 2. The Data Extracted from the User's Play Log

| Information Types | Extracted Factors |
|---|---|
| User's Action | Play Time of Today, The Ranking of Today, Stage Cleared of Today, Character obtained of Today |
| User's Record | The Most Play Time, The Highest Ranking, The Highest stage cleared, Total Character obtained |
| User's Change | Change rate of the Play Time, Change rate of the Ranking, Change rate of the Stage, Change rate of the Character |
| User's Change of Achievement | Change rate of the Play Time, Change rate of the Ranking, Change rate of the Stage, Change rate of the Character |

In order to achieve optimal performance, it is necessary to tune hyperparameters. Bayesian optimization means that the extraction range is re-adjusted based on the previously extracted evaluation results and is efficiently executed without repeating the extraction at random [11]. This method has the advantage of very good performance over time. We adopt Bayesian optimization for hyperparameter tuning.

## 5 RESULTS

Figure 2 shows the training results of the neural network used in the experiment and validation results with data set. Overall, the accuracy is slightly lower, but we observed no signs of serious overfitting.

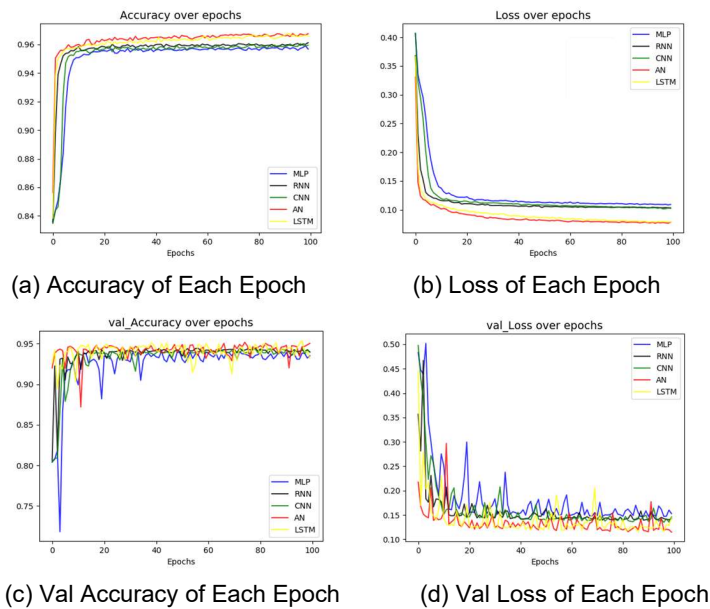

(a) Accuracy of Each Epoch  (b) Loss of Each Epoch

(c) Val Accuracy of Each Epoch  (d) Val Loss of Each Epoch

Figure 2. Accuracy and Loss Curves of Neural Networks

Figure 3 shows the results obtained using the regression prediction algorithm for churn prediction. Most algorithms scored higher when using Churn-vector. In the model predicting the churn date, the result was 0.88 for LASSO, 0.72 for SVM, 0.78 for DT, 0.82 for RF, 0.38 for GBM, 0.84 for MLP, 0.82 for CNN, 0.83 for RNN, 0.85 for LSTM, and 0.88 for AN. The models with the best prediction result are LASSO and AN. In the case of LASSO, the $R^2$ score appears to be high because the $R^2$ score depends on the Mean Squared Error (MSE) value and LASSO's mechanism performs machine learning in a way that reduces the MSE.

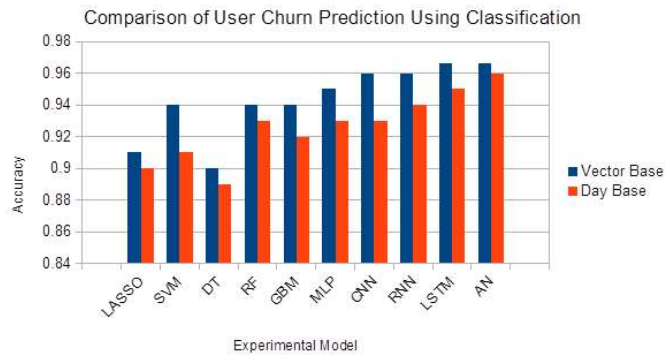

Figure 3. Churn Prediction Using Regression

On the other hand, in the model to predict the churn-vector, the result is 0.88 for LASSO, 0.82 for SVM, 0.90 for DT, 0.91 for RF, 0.70 for GBM, 0.94 for MLP, 0.93 for CNN, 0.94 for RNN, 0.95 for LSTM, and 0.95 for AN. In most cases, the neural network-based algorithms generate higher accuracy than the other approaches. The model with the best prediction result is AN.

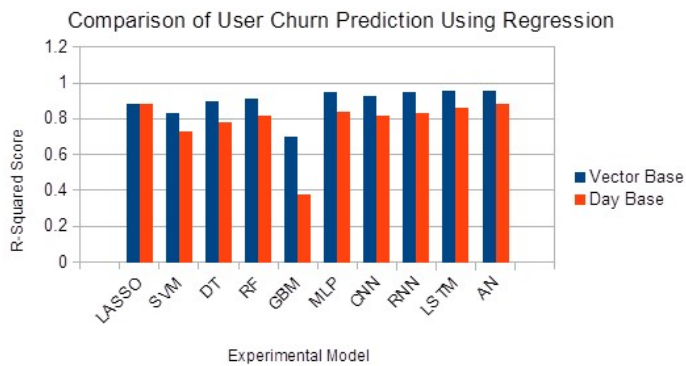

Figure 4. Churn Prediction Using Classification

Figure 4 shows the results obtained using the classification algorithm for churn prediction. Most algorithms scored also higher when using Churn-vector. In the model predicting the churn date, the result was 0.90 for LASSO, 0.914 for SVM, 0.89 for DT, 0.93 for RF, 0.92 for GBM, 0.931 for MLP, 0.933 for CNN, 0.942 for RNN, 0.951 for LSTM, and 0.953 for AN. On the other hand, in the model predicting the churn-vector, the result was 0.91 for LASSO, 0.94 for SVM, 0.90 for DT, 0.94 for RF, 0.94 for GBM, 0.954 for MLP, 0.956 for CNN, 0.957 for RNN, 0.963 for LSTM, and 0.966 for AN. In most cases, the neural network algorithm showed excellent results. The model with the best prediction result was AN.

The experimental results shown in Figures 3 and 4 lead to two conclusions. First, churn vector-based prediction enhances the accuracy of the prediction. Second, AN in neural networks is evaluated as the most accuracy model. We also evaluated the accuracy on a Differential Neural Computer (DNC) -- memory-augmented neural network [12] –- with churn vectors. The DNC is not performed better than other network networks because our hyperparameter tuning does not fit in DNC and DNC's learning speed is too slow with vector-based prediction.

## 6  CONCLUSION

We have seen through a series of explorations that churn vectors can improve performance in predicting churn in mobile games. The contribution of this work shows how neural networks impact the accuracy of churn prediction. It was confirmed that when using an attention network algorithm with the churn vectors, relatively high accuracy is obtained.

Even if this result has been evaluated with the data based on mobile games, many subscription-based businesses such as Netflix, or Amazon Prime, which are sensitive at the risk of churning are likely to adopt our research to minimize customer attrition. Based on our approach, such business could engage their customers with special promotion instead of losing them.